\documentclass[letterpaper, 10 pt, conference]{ieeeconf}  

\IEEEoverridecommandlockouts                              
\usepackage{graphics} 
\usepackage{amsmath} 
\usepackage{amssymb}  
\usepackage{tikz}
\usepackage{import}
\usepackage{pgfplots}
\usepackage{graphicx}
\graphicspath{{./}}
\usepackage{todonotes}
\usepackage[utf8]{inputenc}
\usepackage{array}
\usepackage{booktabs}

\usepackage{url}
\usepackage{algorithmicx}
\usepackage{algorithm}
\usepackage{algpseudocode}

\title{\LARGE \bf
High-performance Racing on Unmapped Tracks using Local Maps
}

\author{Benjamin David Evans$^{1}$,  Hendrik Willem Jordaan$^{1}$ and Herman Arnold Engelbrecht$^{1}$
\thanks{$^{1}$Electrical and Electronic Engineering Department,
        Stellenbosch University, Stellenbosch, 7600, South Africa. 
        {\tt\small bdevans@sun.ac.za};
        {\tt\small hebrect@sun.ac.za};
        {\tt\small wjordaan@sun.ac.za}%
        }
}

\begin{document}

\maketitle
\thispagestyle{empty}
\pagestyle{empty}

\begin{abstract}

Map-based methods for autonomous racing estimate the vehicle's location, which is used to follow a high-level plan.
While map-based optimisation methods demonstrate high-performance results, they are limited by requiring a map of the environment.
In contrast, mapless methods can operate in unmapped contexts since they directly process raw sensor data (often LiDAR) to calculate commands.
However, a major limitation in mapless methods is poor performance due to a lack of optimisation.
In response, we propose the local map framework that uses easily extractable, low-level features to build local maps of the visible region that form the input to optimisation-based controllers.
Our local map generation extracts the visible racetrack boundaries and calculates a centreline and track widths used for planning.
We evaluate our method for simulated F1Tenth autonomous racing using a two-stage trajectory optimisation and tracking strategy and a model predictive controller.
Our method achieves lap times that are 8.8\% faster than the Follow-The-Gap method and 3.22\% faster than end-to-end neural networks due to the optimisation resulting in a faster speed profile.
The local map planner is 3.28\% slower than global methods that have access to an entire map of the track that can be used for planning.
Critically, our approach enables high-speed autonomous racing on unmapped tracks, achieving performance similar to global methods without requiring a track map.

\end{abstract}

\section{Introduction}

Methods for autonomous vehicles can be grouped into map-based or mapless methods.
Classical map-based methods use a perception, planning, and control stack to estimate the vehicle's pose, calculate an optimal trajectory and then track it \cite{wischnewski2022indy}.
Map-based methods are limited to contexts where maps that can be used for localisation exist.
In contrast, mapless methods do not require a map since they calculate control commands directly from the incoming sensor measurements.
Current mapless methods of reactive algorithms \cite{sezer2012novel} and end-to-end neural networks \cite{Brunnbauer2022LatentRacing} achieve poor performance and low completion rates.
In response, this paper addresses the problem of high-performance racing on unmapped tracks.

Map-based methods are limited to contexts where an accurate map has been built, and localisation is available.
Autonomous vehicles should be able to operate in contexts where maps are unavailable or where the environment has changed since the map was built.
For example, self-driving cars must operate in GPS-denied environments where localisation on a map is not possible \cite{mohamed2019survey}.
Mapless vehicle control is difficult due to the challenges in directly interpreting sensor data such as LiDAR scans or camera images.
This has resulted in most optimisation-based planning methods focusing on situations where prebuilt maps are available \cite{Betz2022AutonomousRacing}.

\begin{figure}[t]
    \centering
    \includegraphics[width=\linewidth]{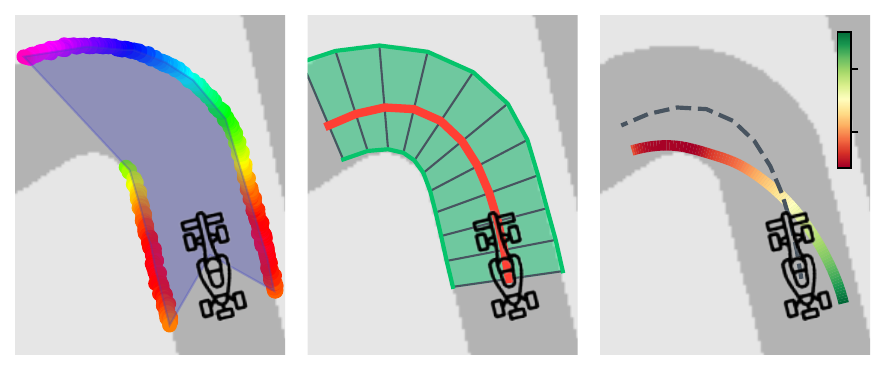}
    \caption{Local map racing pipeline; (1) receive LiDAR scan, (2) extract LocalMap, (3) calculate optimal trajectory.}
    \label{fig:local_trajectory_generation}
\end{figure}

In this paper, we propose using easily extractable, low-level features to build local maps that can be used for optimisation-based planning and enable high-performance control in unmapped environments.
We present this solution in the context of simulated F1Tenth autonomous racing on unmapped tracks \cite{o2020f1tenth}.
Fig. \ref{fig:local_trajectory_generation} shows how our approach uses the LiDAR scan to construct a representation of the visible environment that is used for optimisation-based racing.
We evaluate our method with a two-layer optimisation and tracking strategy \cite{Heilmeier2020MinimumCar}, and a single-layer Model Predictive Contouring Control (MPCC) algorithm \cite{liniger2015optimization}.
We compare our methods to state-of-the-art mapless approaches of end-to-end neural networks \cite{evans2023high} and the Follow-The-Gap (FTG) algorithm \cite{sezer2012novel}.

\section{Literature Study}

\begin{figure}[h] 
    \centering
    \includegraphics[width=\linewidth]{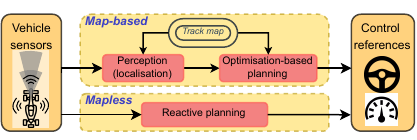}
    \caption{Map-based planners using a perception and optimisation-based strategy compared with mapless methods.}
    \label{fig:classical_racing}
\end{figure}

Fig. \ref{fig:classical_racing} shows how autonomous racing methods can be split into classical methods, which use a perception and optimisation-based planning approach, and mapless methods, which use reactive control algorithms.

\begin{figure*}[h]
    \centering
    \vspace{3mm}
    \includegraphics[width=0.9\textwidth]{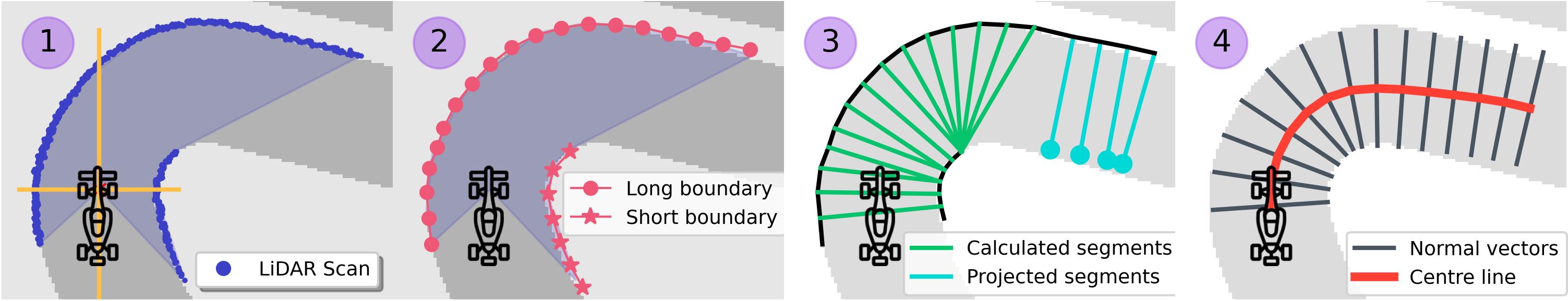}
    \caption{Local map extraction: the track boundaries are identified and used to calculate a centre line and normal vectors of the visible region.}    
    \label{fig:extraction_square}
\end{figure*}

\subsection{Classical Racing}

Classical racing methods use a perception, planning and control pipeline to move the vehicle around the track as quickly as possible.
Racing tracks are typically mapped using a Simultaneous Localisation And Mapping (SLAM) algorithm that requires a slow pass around the track to build a map \cite{nobis2019autonomous}.
During the drive through the map, the estimated odometry is fused with incoming LiDAR scans and/or camera images and used to build a map \cite{andresen2020accurate}.
Localisation for autonomous racing is most commonly done with a scan-matching estimation algorithm that uses incoming odometry and sensor measurements to estimate the vehicle's location \cite{stahl2019ros, WalshCDDTLocalization}.
While localisation methods, such as particle filters, provide robust, accurate localisation, they are inherently limited by requiring a map of the race track. 
That makes them unsuitable for racing in unmapped or dynamically changing environments.

Classical racing typically uses a trajectory optimisation strategy to generate planning references of speed and steering angle.
It is common to use a two-stage planning approach of calculating an optimal trajectory \cite{Heilmeier2020MinimumCar} and then following it with a path-following algorithm, such as pure pursuit \cite{o2020tunercar, becker2022model}.
Another approach is to use a single-layer model predictive control (MPC) to directly optimise control commands that result in an optimal trajectory for a receding horizon \cite{liniger2015optimization, cataffo2022nonlinear}.
While optimal control strategies are effective for high-performance racing, they are limited by requiring a map of the track and the vehicle's current location.

\subsection{Mapless Racing}

Mapless methods have been widely studied in the navigation literature, where the general problem is to move a holonomic robot from one point to another \cite{yan2020mapless}.
One of these methods, Follow-The-Gap \cite{sezer2012novel}, has been adapted (also known as the disparity extender algorithm) to autonomous racing where it has won a race \cite{nathan2023disparityExtender}.
However, these methods have no inherent model for speed selection and thus cannot operate the vehicle near the performance limits. 

Another upcoming method is end-to-end deep learning, which uses a neural network to map raw sensor data directly to control commands.
Bosello et al. \cite{Bosello2022TrainRaces} demonstrated that these methods can generalise to unseen race tracks. 
Current shortcomings in these methods are high-crash rates, even in simulation \cite{hamilton2022zero}, the simulation-to-reality gap \cite{murdoch2023partial}, and jerky action selection \cite{evans2023high}.

We aim to retain the high-performance nature of optimisation-based racing without requiring a global map.
We do this by building a local map of the visible region that can be used for classical planning methods.


\section{Methodology}

We present the local map framework in the context of autonomous racing, where vehicles have a 2D LiDAR sensor for input and must select speed and steering actions that move the vehicle around the track as quickly as possible.

\subsection{Local Map Extraction}

The local map extraction uses the 2D LiDAR scan to build a set of centre line points and track widths that can be used for planning.
Fig. \ref{fig:extraction_square} graphically illustrates this process, and Algorithm \ref{alg:lm_algorithm} provides pseudo code for the implementation.

\begin{algorithm}[h]
    \caption{Local map extraction algorithm}
    \begin{algorithmic}[1]
        \State Receive LiDAR scan $\textbf{l}$
        \State  $Z = \{[l_i \cos(\theta_i),~ l_i \sin(\theta_i)]\}$ \Comment{Scan to Cartesian points}
        \State $Z \rightarrow B^\text{long},~ B^\text{short}$ \Comment{Identify boundaries}
        \State Resample boundaries with $n^\text{long}$ and $n^\text{short}$ points
        \State $S = \varnothing$ \Comment{Initialise segment list}
        \For{$i = 0, ~n^\text{long}$}
            \State \text{distances} = $\{| \textbf{b}_i^\text{long} - \textbf{b}_j^\text{short}|_2 ~~ \forall ~~ j \in [0, ~n^\text{short}]\}$
            \State $k_i = \text{argmin} (\text{distances})$ 
            \If{$| \textbf{b}_i^\text{long} - \textbf{b}_j^\text{short}|_2 < w_\text{max}$}
                \State Append [$\textbf{b}_i^\text{long},~ \textbf{b}^\text{short}_{k_i}$] to $S$
            \Else 
                \State $R^\text{long} \leftarrow B^\text{long}[i:n^\text{long}]$ \Comment{Remaining long line}
                \State $R^\text{short} = R^\text{long} + \texttt{normals}(R^\text{long}) \times w_\text{track}$ 
                \State Append $[R^\text{long},~ R^\text{short}]$ to $S$
                \State break out of for loop
            \EndIf
        \EndFor
        \State Use list $S$ to calculate centre line and track widths
    \end{algorithmic}
    \label{alg:lm_algorithm}
\end{algorithm}

Fig \ref{fig:extraction_square} (1), shows the vehicle's LiDAR scan as a set of points, defined by distances at set angles from the LiDAR scanner.
Line 2 of the pseudo-code describes how they are converted to Cartesian points, $Z$, in the vehicle's inertial frame, by multiplying by the corresponding angles $\theta_i$.
The LiDAR's number of beams $N_\text{beams}$ and field-of-view angle $A$ are used to calculate the set of angles as, $$\{-A/2 + n * A/N_\text{beams} | n \in [0,1,..., N_\text{beams}]\}$$.
Fig \ref{fig:extraction_square} (2) shows how the set of points is split into left and right track boundary lines.
Pseudocode lines 3 and 4 explain that the boundaries are treated in the categories of \textit{long} and \textit{short} and resampled to have equidistant points.

Lines 6-10 describe the progress to generate the green lines in Fig \ref{fig:extraction_square} (3).
For each point in the long boundary $\textbf{b}_i^\text{long}$, the distances to all the points on the short boundary are calculated (line 7).
The nearest point $\textbf{b}^\text{short}_{k_i}$ is found by taking the \textit{argmin} of the distance array (line 8).
If the distance between the points is smaller than the maximum track width $w_\text{max}$, then the segment is added to the segment list $S$ (line 10).
For the section where both boundaries are visible, this process finds track segments approximately normal to the track direction.

Lines 11-16 explain how the centre line is extended where only the longer boundary is visible.
Line 12 identifies $R^\text{long}$ as the long line section from the last segment added until the end of the line.
The \texttt{normals} function returns the normal vectors indicating the direction across the track to the other boundary.
The estimated short boundary $R^\text{short}$ is then calculated in line 13 by adding the long boundary and the normal vectors multiplied by the track width $w_\text{track}$.
The estimated segments $R^\text{long}$ and $R^\text{short}$, shown as turquoise lines in Fig. \ref{fig:extraction_square} (3) are added to the segment list $S$.

\begin{figure}[h]
    \centering
    \begin{minipage}{0.48\linewidth}
        \includegraphics[width=\textwidth]{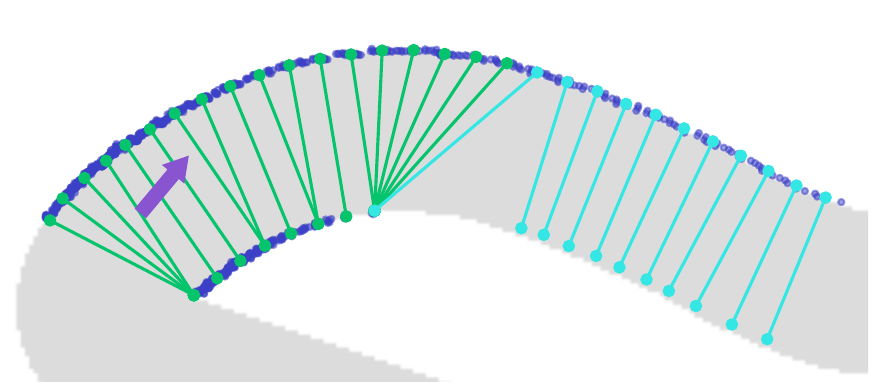}
    \end{minipage}
    \hfill
    \begin{minipage}{0.48\linewidth}
        \includegraphics[angle=270, width=\textwidth]{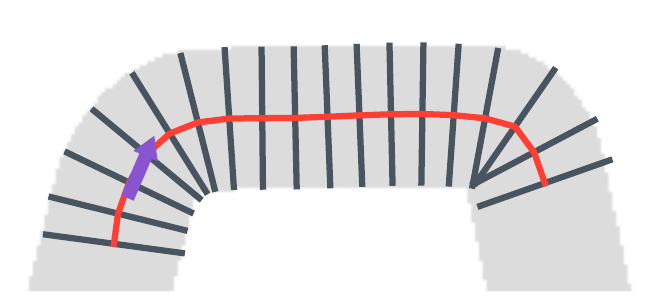}
    \end{minipage}
    \caption{Local map segment extraction (left) and a final local map (right) for segments of the AUT track. The purple arrow represents the car's pose.}
    \label{fig:local_map_extraction}
\end{figure}

The left image in Fig. \ref{fig:local_map_extraction} shows an example segment extraction, with the blue dots representing the LiDAR scan, the green lines representing the calculated segments, and the turquoise lines representing the projected segments.
The green line segments ensure that the centre of the track is calculated since both boundaries are known.
The turquoise line segments assume the track has a constant track width and that the borders are parallel to each other.
The segment list is used to find the track centre line by adding finding the points in between the two boundaries.
The line is then resampled to have equidistant points, and the track widths are calculated.
The right image in Fig. \ref{fig:local_map_extraction} shows the resulting local map centre line and normal vectors that is returned and used for planning.

\subsection{Optimisation Planners} \label{subsec:opti_strategies}

We implement two standard planning techniques: a two-stage optimisation and tracking approach and a single-layer model predictive controller.
These planning strategies are used with either global maps of the entire track and localisation or a local map.

\subsubsection{Two-stage Optimisation Planner}

\begin{figure}[h]
    \centering
    \includegraphics[width=0.8\linewidth]{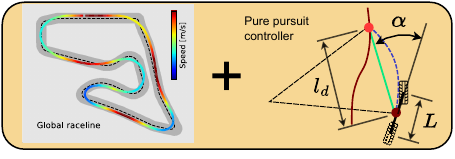}
    \caption{The two-stage planner generates an optimal trajectory that is tracked with a pure pursuit controller}
    \label{fig:two_stage}
\end{figure}

Fig. \ref{fig:two_stage} shows how the two-stage planner generates an optimal trajectory that is tracked using the pure pursuit algorithm.
We use the optimisation approach presented by Heilmeier et al. \cite{Heilmeier2020MinimumCar}, generating a minimum curvature path and then a minimum time speed profile.
The optimisation generates a path by minimising the curvature, which is defined as the rate of change in the heading.
A forward-backwards solver calculates the speed profile by selecting the fastest speed at each point that keeps the vehicle within the friction limits and is dynamically reachable.

The pure pursuit formula \cite{coulter1992implementation} (shown on the right in Fig. \ref{fig:two_stage}) calculates a steering angle that tracks an upcoming waypoint that is lookahead distance $l_\text{d}$ away, at an angle of $\alpha$ as,
\begin{equation}
    \delta = \arctan \bigg( \frac{L \sin (\alpha)}{l_\text{d}} \bigg).
\end{equation}
The lookahead distance is selected based on the speed as,
$l_\text{d} = l_c + l_s \times V$
where $l_c$ and $l_s$ are hyperparameters and $V$ is the vehicle speed.

\subsubsection{Model Predictive Contouring Control}

We adapt the Model Predictive Contouring Control (MPCC) algorithm presented by Liniger et al. \cite{liniger2015optimization} to F1Tenth racing.
The optimisation represents the vehicle state as position and heading and the inputs as steering and speed for a finite number of steps.
Successive states are constrained to the vehicle dynamics, where a kinematic bicycle model is used to update the states based on the control inputs.
Additionally, the trajectory is constrained to lie within the track boundaries.

\begin{figure}[h]
    \centering
    \includegraphics[width=0.9\linewidth]{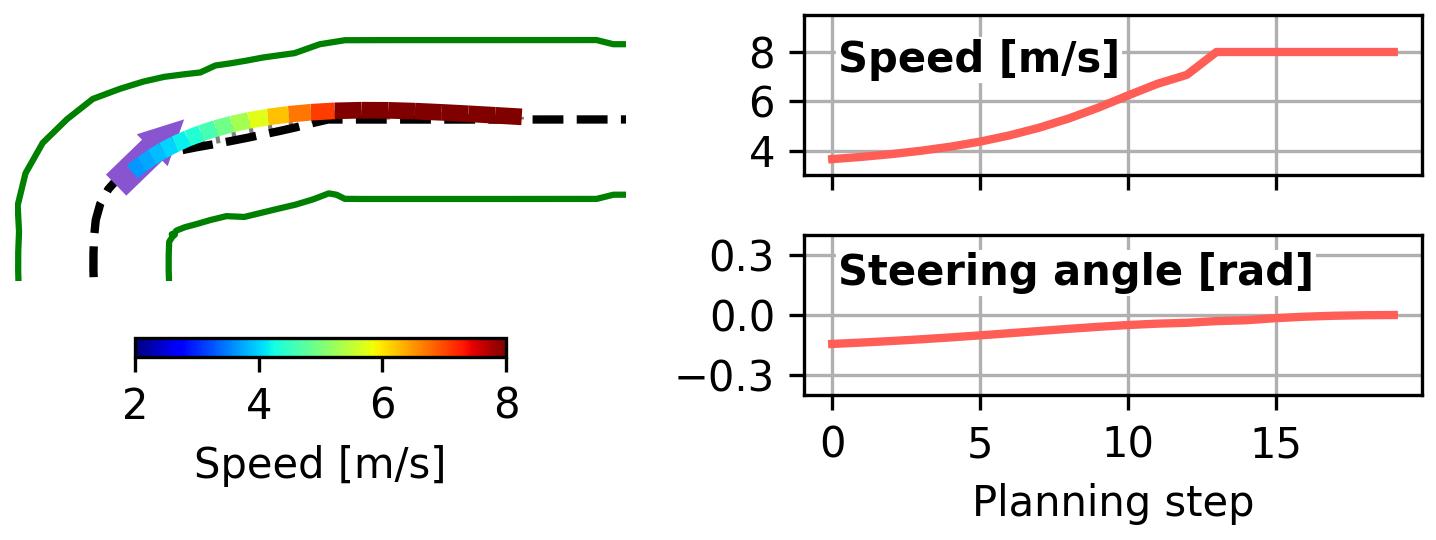}
    \caption{The MPCC algorithm plans a receding horizon trajectory of speed and steering angle references that maximises centre-line progress.}
    \label{fig:mpcc_description}
\end{figure}

Fig. \ref{fig:mpcc_description} shows how the MPCC algorithm plans a finite trajectory by selecting speed and steering angle references.
Since the nearest point on the reference path cannot be directly found, an approximate point on the centre line is used and added as an additional state to the optimisation variables.
A lag error penalises the difference between the approximate centre line point and the true point along the trajectory.
A contouring error encourages the trajectory to track the centre line.
A progress objective promotes progress along the path, and a control regulation term encourages smooth control actions.

\section{Evaluation}

\subsection{Methodology}

The evaluation compares local map planners with global map planners and mapless methods.
We implement the two optimisation strategies presented in Section \ref{subsec:opti_strategies} with the global and local planners.
We use baseline mapless methods of end-to-end neural network controllers \cite{evans2023high} and the FTG method \cite{sezer2012novel}.
We train end-to-end agents with the SAC and TD3 reinforcement learning algorithms using two hidden layers of 100 neurons, the trajectory-aided learning reward and 60,000 training steps on the GBR map.


\begin{figure}[h]
    \centering
    \begin{tabular}{ccc}
      \includegraphics[width=1.4cm]{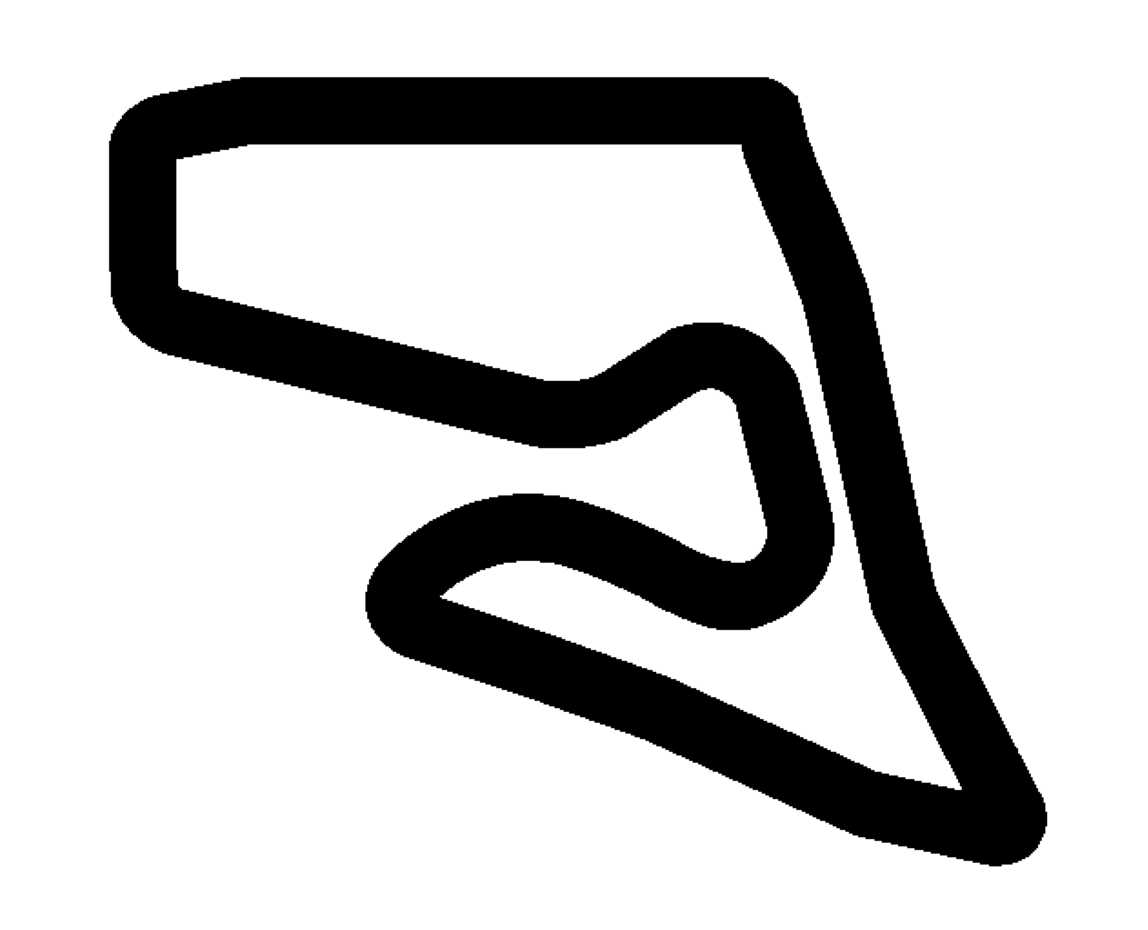}  & \includegraphics[width=2.3cm]{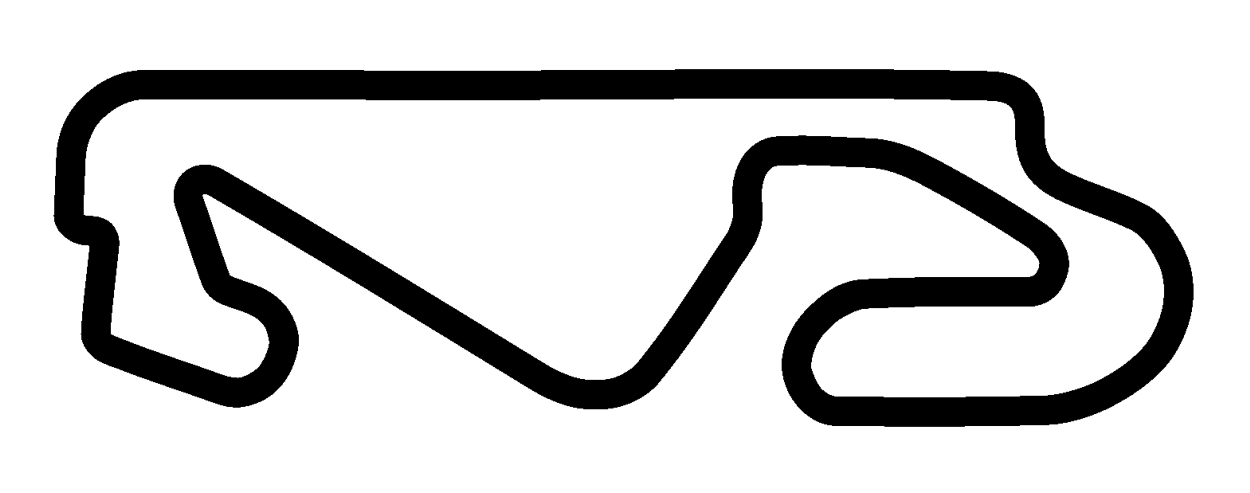}   & \includegraphics[width=1.9cm]{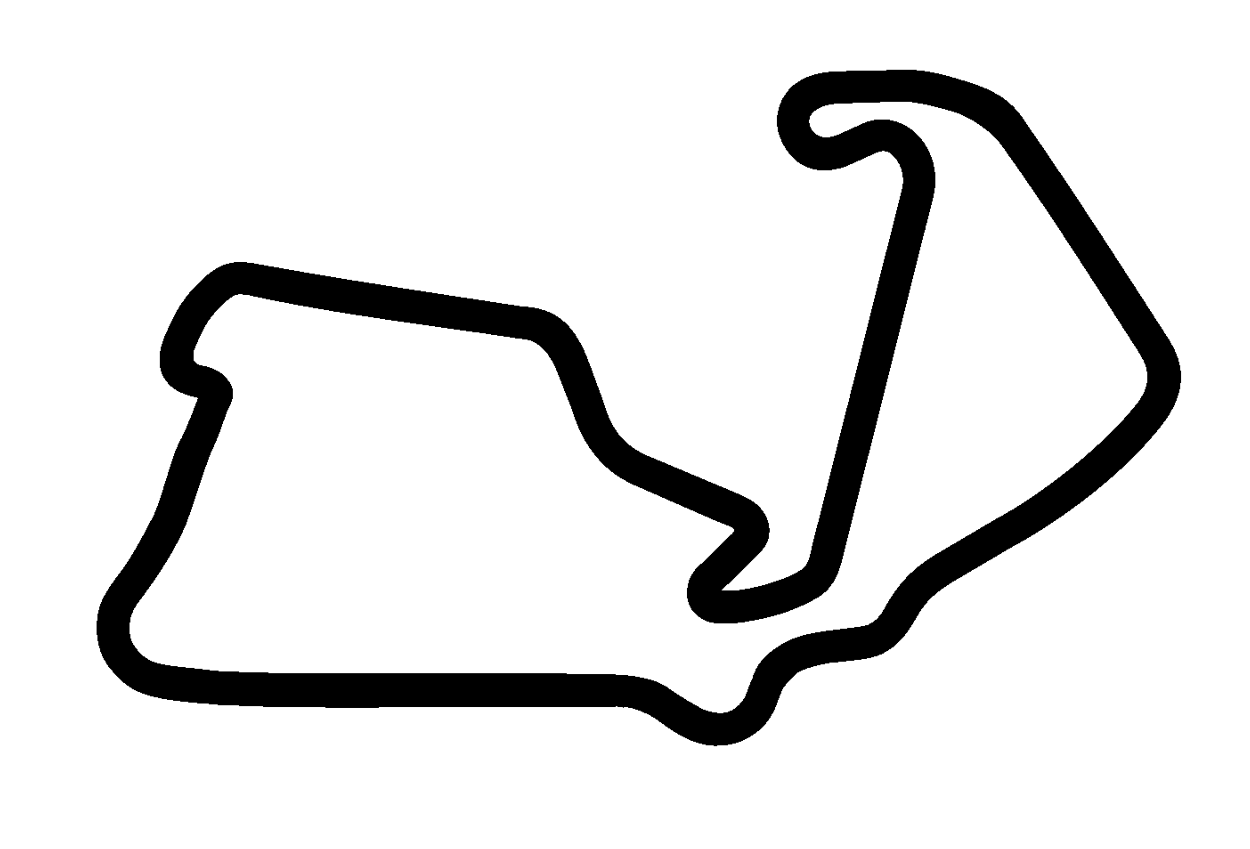} \\
    \end{tabular}
    \caption{The AUT, ESP, and GBR (left to right) track maps.}
    \vspace{-2mm}
    \label{fig:map_description}
\end{figure}

We evaluate our method for F1Tenth autonomous racing using the simulator presented by O'Kelly et al. \cite{o2020f1tenth}.
The vehicle is represented using the single-track bicycle model, and the dynamics equations are updated at 100 Hz.
All the planners are run at 25 Hz, and at each planning step, a steering angle and linear speed must be selected to control the car.
Fig. \ref{fig:map_description} shows the AUT, ESP and GBR maps used for simulation testing.
We conduct the following tests:
\begin{enumerate}
    \item \textbf{Local map extraction:} We investigate our pipeline's ability to extract local maps
    \item \textbf{Racing performance comparison:} We compare the lap times and speed profiles of the planners 
    \item \textbf{Computational requirements:} We measure the computation of each algorithm
\end{enumerate}


\subsection{Local Map Extraction}

We investigate the local map extraction pipeline that uses the incoming 2D LiDAR scan to generate a local map.
We measure the length of the centre lines of the extracted local maps for each track by using the pure pursuit algorithm to track the centre line at a low, constant speed.

\begin{figure}[h]
    \centering
    \vspace{3mm}
    \includegraphics[width=0.9\linewidth]{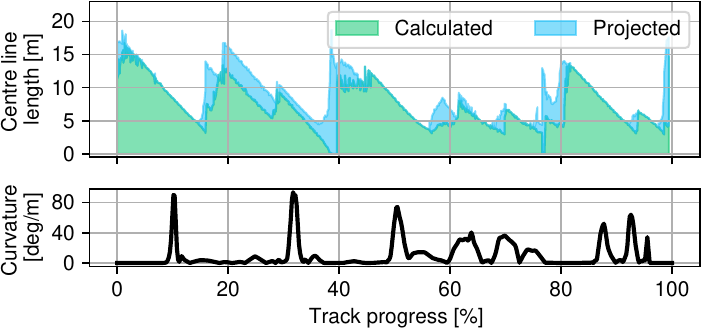}
    \caption{Calculated and projected local map lengths compared to curvature.}
    \vspace{-3mm}
    \label{fig:track_length_plot}
\end{figure}

Fig. \ref{fig:track_length_plot} shows the calculated and projected local map lengths compared to curvature for a lap on the AUT track.
The centre line lengths show a pattern of downward slopes followed by sharp rises.
The curvature graph shows that the sharp rises appear after a curvature spike, indicating that the sharp rise is the straight that is suddenly visible after turning a corner.

\begin{table}[h]
    \centering
    \renewcommand{\arraystretch}{1.4}
        \begin{tabular}{w{l}{2.1cm} >{\centering\arraybackslash}m{1.6cm} >{\centering\arraybackslash}m{1.6cm} >{\centering\arraybackslash}m{1.6cm}}
            \toprule
             Statistic & AUT & ESP & GBR \\
            \midrule
             Mean  $\pm$ Std. dev. & 11.05 $\pm$ 4.52 & 11.10 $\pm$ 4.76 & 11.03 $\pm$ 4.77 \\
             Min, Max & 3.44,  21.35 & 2.57, 25.96 & 2.64, 25.96 \\
            \bottomrule
        \end{tabular}
    \caption{The mean, standard deviation, minimum and maximum local map lengths for the AUT, ESP and GBR tracks.}
    \label{tab:centre_line_table}
\end{table}
\vspace{-5mm}

Table \ref{tab:centre_line_table} shows that the maps have a mean centre line length of around 11 m, with a standard deviation of around 4.5 m. 
The local map lengths range between 2.57 m and 25.96 m.
We conclude that the centre line can robustly be extracted and used for planning for the entirety of the track.

\subsection{Racing Performance Comparison}

We investigate the racing performance of the global and local optimisation planners, the Follow-The-Gap algorithm and the SAC and TD3 end-to-end agents.
For each planner, five test laps are run with random start positions, and the average times from completed laps are presented.

\begin{figure}[h]
    \centering
    \includegraphics[width=\linewidth]{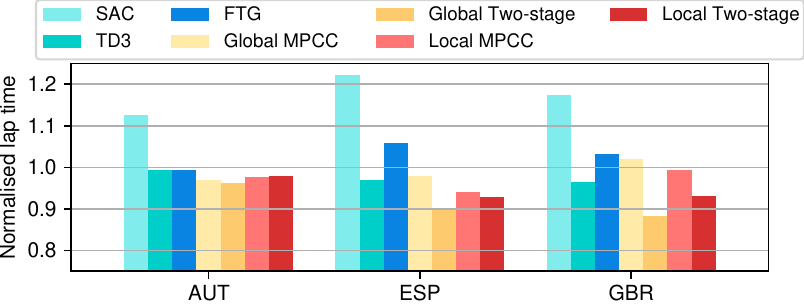}
    \caption{Normalised lap times from the SAC and TD3 agents, FTG method, and global and local MPCC and two-stage planners.}
    \label{fig:normalised_lap_times}
\end{figure}

Fig. \ref{fig:normalised_lap_times} shows a bar graph of the lap times for each planner, normalised by dividing by the mean lap time for each track.
The end-to-end agents and Follow-The-Gap method have the slowest times on all the maps. 
The global and local planners achieve fast times, with the two-stage planners outperforming the MPCC planners.
Therefore, we study the performance of the two-stage planners, TD3 agent and Follow-The-Gap method in detail.

\begin{table}[h]
    \centering
    \renewcommand{\arraystretch}{1.4}
    \begin{tabular}{w{l}{0.9cm} >{\centering\arraybackslash}m{1.5cm} >{\centering\arraybackslash}m{1.7cm} >{\centering\arraybackslash}m{1.6cm} >{\centering\arraybackslash}m{0.8cm}}
        \toprule
        \textbf{Map} & \textbf{TD3} & \textbf{FTG} & \textbf{Global} & \textbf{Local} \\
        \midrule
        \textbf{AUT} & 19.13 (1.5\%) & 19.11 (1.4\%) & 18.52 (-1.8\%) & 18.85  \\
        \textbf{ESP} & 41.95 (4.5\%) & 45.77 (14.1\%) & 38.97 (-2.9\%) & 40.13  \\
        \textbf{GBR} & 36.71 (3.6\%) & 39.31 (11.0\%) & 33.58 (-5.2\%) & 35.42  \\
        \midrule
        Mean & 3.22\% & 8.80\% & -3.28\% & - \\
        \bottomrule
    \end{tabular}
    \caption{Lap time in seconds (\% difference from the local planner) for the TD3, FTG, and global and local two-stage planners.}
    \label{tab:lap_time_table}
\end{table}
\vspace{-5mm}

\begin{figure}[h]
    \centering
    \includegraphics[height=0.35\linewidth]{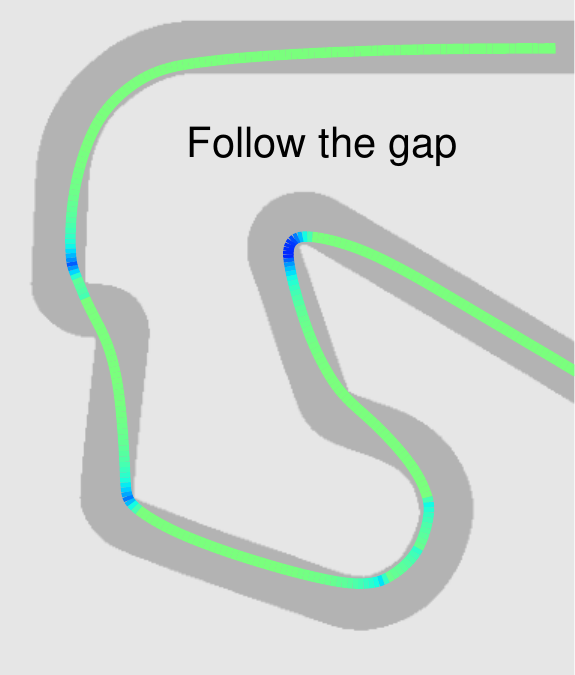}
    \hfill
    \includegraphics[height=0.35\linewidth]{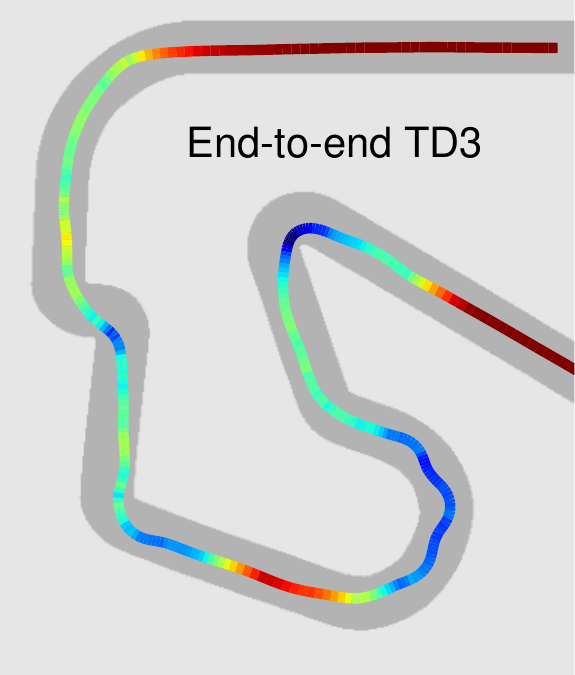}
    \hfill
    \includegraphics[height=0.35\linewidth]{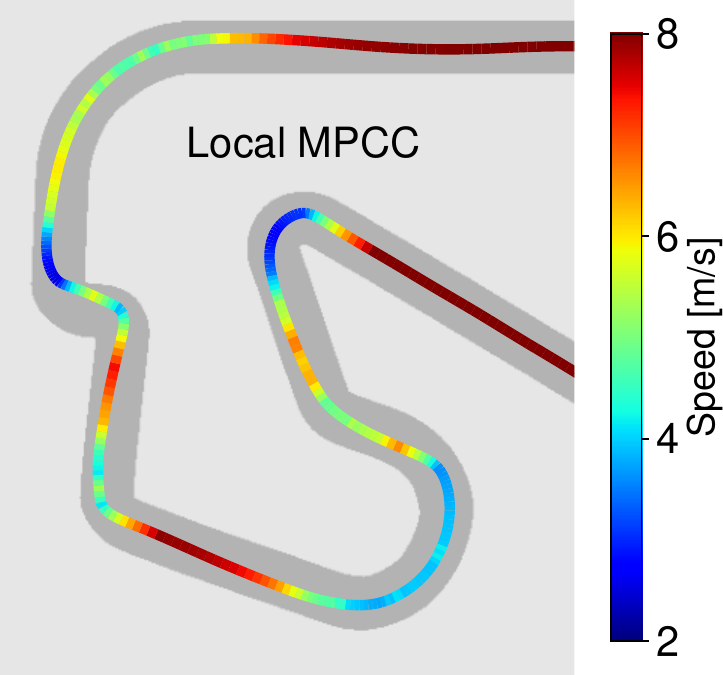}
    \caption{Trajectory segments on the ESP map of the follow the gap, end-to-end and local MPCC planners. }
    \vspace{-2mm}
    \label{fig:trajectory_comparison}
\end{figure}

Table \ref{tab:lap_time_table} shows the lap times and percentage difference from the local planner.
The local map planner outperforms the TD3 agent by 3.22\% and the Follow-The-Gap planner by 8.8\%.
We investigate this result by plotting the trajectory segments on the ESP map in Fig. \ref{fig:trajectory_comparison}.
The FTG planner takes a short, smooth path but has poor speed selection due to not using a model of the track.
The end-to-end planner selects a more appropriate speed profile, but the trajectory is not smooth, and consequently, the vehicle selects conservative speeds.
The local MPCC planning speeds up in the straights and slows down in the corners while selecting a smooth path.
This fast performance outperforms other mapless methods.

\begin{figure}[h]
    \centering
    \includegraphics[height=0.35\linewidth]{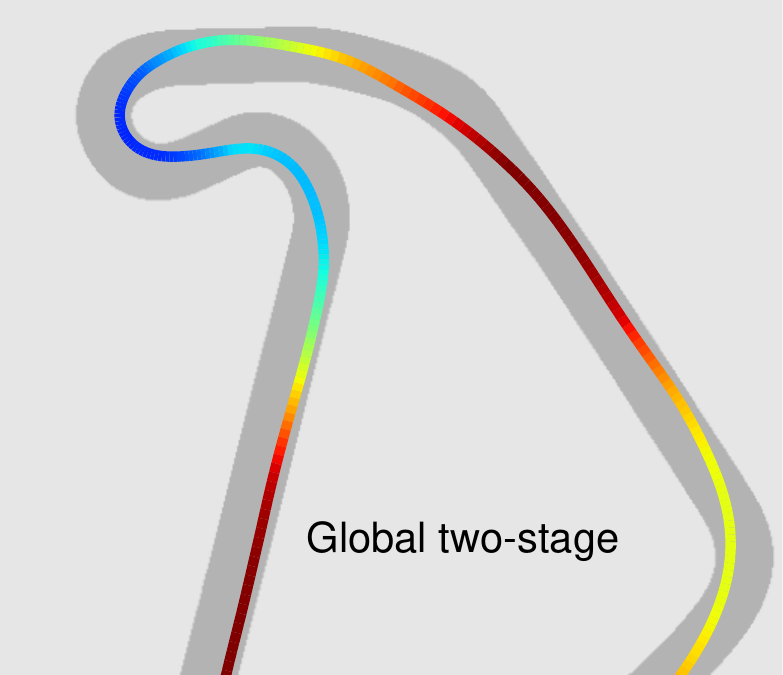}
    \includegraphics[height=0.35\linewidth]{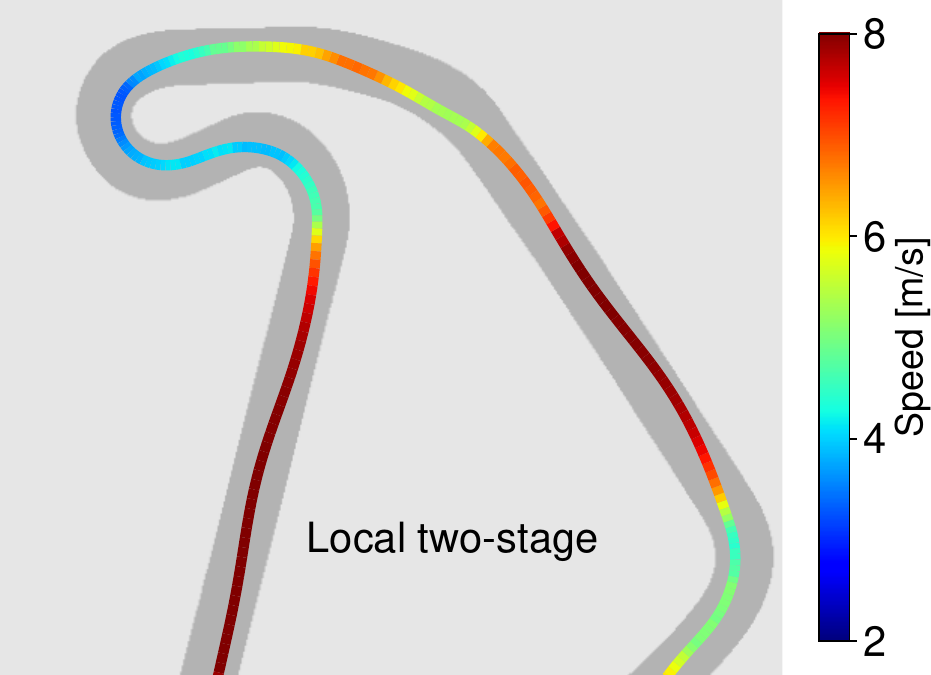}
    \caption{Global and local two-stage planner trajectories on the GBR map.}
    \vspace{-2mm}
    \label{fig:global-vs_local_gbr}
\end{figure}

Table \ref{tab:lap_time_table} shows that the global two-stage planner achieves faster lap times on all the maps than the local two-stage planner.
On average, the global planner lap times are 3.28\% faster.
We investigate this by plotting trajectory segments for a portion of the GBR map in Fig. \ref{fig:global-vs_local_gbr}.
The two trajectories show a similar pattern of speeding up to high speeds in the straighter sections and slowing down enough for the corners, with the local planner slowing down more in the turns.

\begin{figure}[h] 
    \centering
    \vspace{3mm}
    \includegraphics[width=0.9\linewidth]{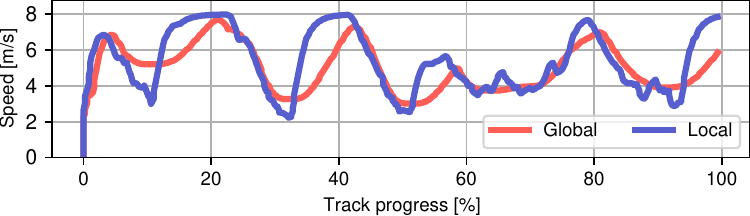}
    \vspace{-2mm}
    \caption{Comparison of speed profiles from global and local two-stage planners on the AUT track.}
    \label{fig:SpeedComparison_aut}
\end{figure}

Fig. \ref{fig:SpeedComparison_aut} shows the speed profiles from the global and local two-stage planners on the AUT track.
The global planner smoothly speeds up and slows down due to its ability to optimise the profile over the entire track.
In contrast, the local planner shows more extreme behaviour of quickly speeding up to the maximum speed and slowing down to low speeds of around 2 m/s in the corners.

\begin{figure}[h]
    \centering
    \includegraphics[height=0.13\linewidth]{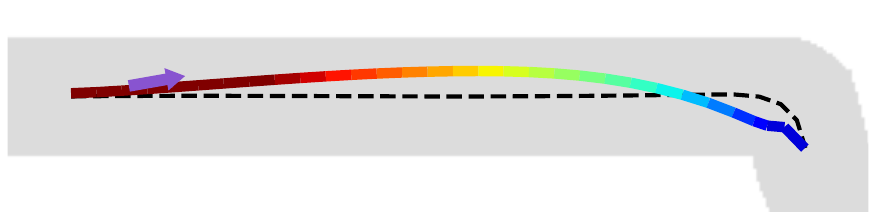}
    \hfill
    \includegraphics[height=0.19\linewidth]{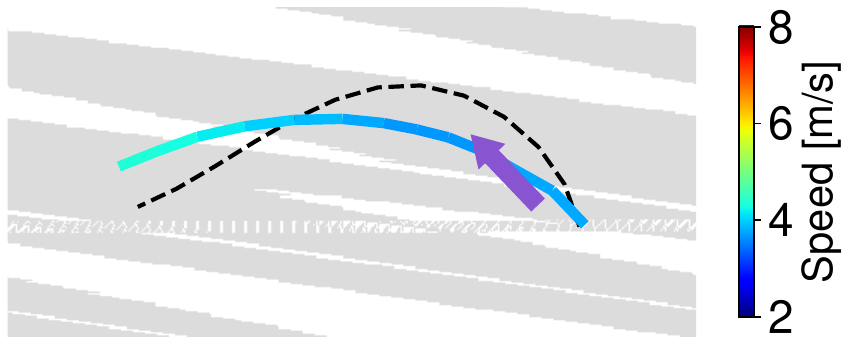}
    \vspace{-2mm}
    \caption{Local racelines generated by the two-stage planner on AUT.}
    \label{fig:local_raceline_gen}
\end{figure}

Fig. \ref{fig:local_raceline_gen} shows two example local racelines with the vehicle location represented by the purple arrow.
These racelines illustrate the limited visible planning horizon, which results in the local planner selecting a conservative trajectory.


Our approach outperforms previous state-of-the-art approaches to autonomous racing of end-to-end neural networks and the Follow-The-Gap method by achieving lower lap times and smoother paths.
The reason for this is that using local map representations of the visible track enables the use of an optimisation strategy.
In comparison to global planning methods, our approach shows only a 3\% performance drop.
The drop is explained by the local map planner's conservatism around corners due to the lack of visibility.
The local map planner selects a speed profile similar to the global planner's while using only the currently visible local map.

\subsection{Computational Requirements}

We investigate the perception and planning computation times of the racing algorithms.
The perception refers to localisation using a particle filter for the global planners and local map generation for the local planners.
All the tests are written in Python, run on an Intel i7-10700 desktop computer running Ubuntu 22.04 and use the cProfile library.

\begin{figure}[h]
    \centering
    \includegraphics[width=\linewidth]{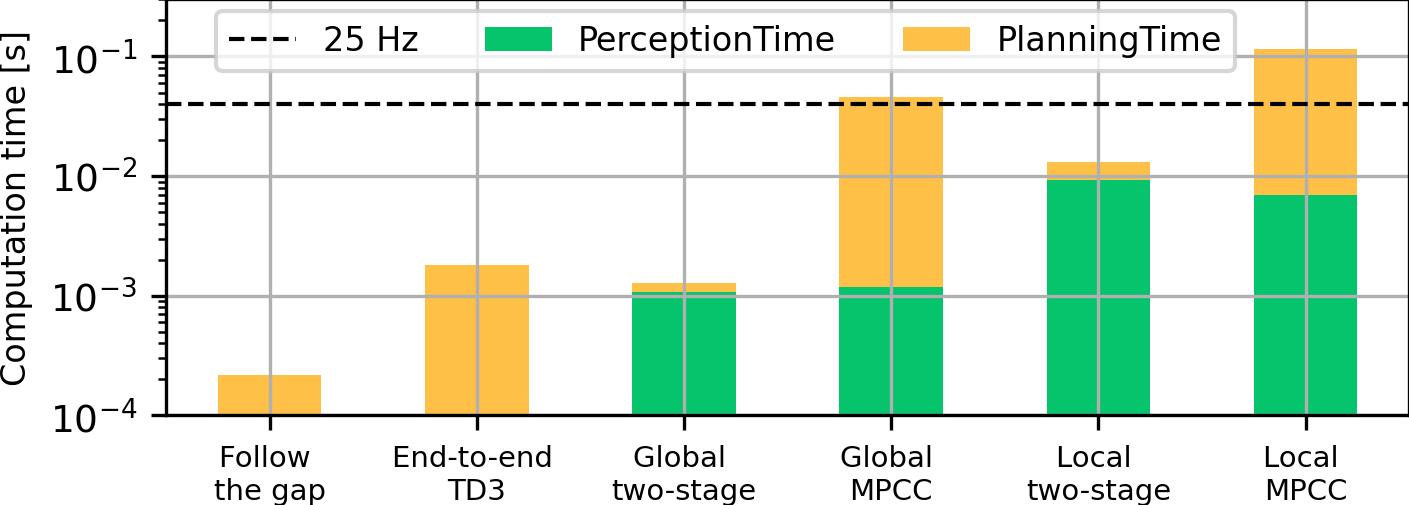}
    \caption{Computation times for perception and planning on ESP.}
    \label{fig:computation_times}
\end{figure}

Fig. \ref{fig:computation_times} shows each planner's average perception and planning computation times.
The Follow-The-Gap algorithm takes around 0.3 ms to plan, followed by the end-to-end agent taking around 1.2 ms.
The global planners have perception (localisation) times of around 1 ms, and the local planners have longer perception (local map generation) times of around 10 ms.
The two-stage planners run quickly, taking only 0.1 ms for the global and 0.3 ms for the local planner.
The dashed black line shows the time required to run the algorithm at 25 Hz (40 ms).
The global and local MPCC algorithms exceed the time constraint for real-time computation.
The local map planner can run in real time if combined with an efficient algorithm.

\section{Conclusion}


This article presented the local map framework for optimisation-based control in unmapped environments.
The LiDAR scan was used to extract a local map of the visible area for F1Tenth autonomous racing.
Our method extracted local maps that could be used for input to an optimisation strategy for planning.
Local maps with an average of 11 m can be extracted from differently shaped racetracks.
Our two-stage local map planner achieves lap times 3.22\% faster than end-to-end agents trained with the TD3 algorithm and 8.8\% faster than the Follow-The-Gap method.
The primary improvement source is the local map planner's access to a vehicle dynamics model that can be optimised.
The comparison with global planning approaches showed an average of 3.28\% slower lap times, resulting from limited planning horizon around corners and thus reduced speed.
Critically, our approach removes the limitation of global approaches in requiring a track map and localisation and, thus, enables high-performance racing on unmapped tracks.

\subsection{Future Work}

\textbf{Image-based Control:}
Image-based control is difficult due to the large variability in images.
Image-based mapping and localisation methods are computationally expensive, and end-to-end methods have not yet demonstrated satisfactory results \cite{cai2021vision}.
However, the local map framework could use an edge detection algorithm to detect track boundaries and build a local map.
Extracting low-level features can be used to enable high-performance image-based control.

\textbf{Local Map Fusion for High-speed SLAM:}
Once a lap of a race track has been completed with the LocalMap planner, the local maps could be fused together to form a global map.
Further, this can be extended to a full racing-optimised version of SLAM that builds a map in real-time by fusing successive local maps and estimating the difference.
This will result in efficient, high-speed racing SLAM.

\textbf{Safe Reinforcement Learning:}
Safe reinforcement learning, where the agent learns onboard a physical vehicle, has previously required a track map to build a kernel of safe states \cite{evans2023safe}.
Previously, safe learning was limited to mapped contexts where localisation was available.
Using local maps would enable a kernel of safe states to be built online, thus enabling safe learning to occur on unmapped tracks.
Safe learning has the potential to achieve high-performance racing since it avoids the simulation-to-reality gap.

\addtolength{\textheight}{-12cm}   

\typeout{}


\end{document}